\pdfoutput=1 

\documentclass[letterpaper, 10 pt, conference]{ieeeconf}  

\IEEEoverridecommandlockouts                              

\usepackage{bm}
\usepackage{cite}
\usepackage{flushend}
\include{preamble}

\title{
\LARGE \bf
High-Power, Flexible, Robust Hand: \\Development of Musculoskeletal Hand Using Machined Springs \\and Realization of Self-Weight Supporting Motion with Humanoid
}

\author{Shogo Makino, Kento Kawaharazuka, Masaya Kawamura, Yuki Asano, Kei Okada, Masayuki Inaba$^{1}$%
\thanks{$^{1}$The authors are with the Department of Mechano-Informatics, The University of Tokyo 7-3-1 Hongo, Bunkyo-ku, 113-8656 Tokyo, Japan
        {\tt\small \{makino, kawaharazuka, kawamura, asano, k-okada, inaba\}@jsk.t.u-tokyo.ac.jp}}%
}

\begin{document}

\setlength{\pdfpageheight}{11in}
\setlength{\pdfpagewidth}{8.5in}

\maketitle
\thispagestyle{empty}
\pagestyle{empty}

\begin{abstract}
Human can not only support their body during standing or walking, but also support them by hand, so that they can dangle a bar and others. But most humanoid robots support their body only in the foot and they use their hand just to manipulate objects because their hands are too weak to support their body. Strong hands are supposed to enable humanoid robots to act in much broader scene.
Therefore, we developed new life-size five-fingered hand that can support the body of life-size humanoid robot. It is tendon-driven and underactuated hand and actuators in forearms produce large gripping force. This hand has flexible joints using machined springs, which can be designed integrally with the attachment. Thus, it has both structural strength and impact resistance in spite of small size. As other characteristics, this hand has force sensors to measure external force and the fingers can be flexed along objects though the number of actuators to flex fingers is less than that of fingers.
We installed the developed hand on musculoskeletal humanoid ``Kengoro'' and achieved two self-weight supporting motions: push-up motion and dangling motion.
\end{abstract}

\section{INTRODUCTION}
Human can not only support their body during standing or walking, but also support them by hand. For example, they can dangle a bar, climb a wall and do push-ups. Furthermore, they can stabilize their body by using hands even when they do the tasks which can be done without using. For example, they can go up the stairs and walk on uneven grounds while gripping handrail. On the other hand, most humanoid robot move just on foot, and they use their hands just to grasp and manipulate objects. In the first place, one of the purpose of developing humanoid robots is to work in disaster site instead of human as DARPA Robotics Challenge were held. Thus, humanoid robots are needed to advance through harsh environment. In such an environment, in most case, humanoid robots cannot always walk normally. In recent year, studies have been conducted on the whole-body motion, in which humanoid robot support their body by four limbs \cite{toro_iros2016, Noda_icra2014}. However, in this study the load which hands support was not so large. One of the reason why humanoid robots cannot support large load by hand is the weakness of their hands. If humanoid robots have enough strong hands to support their body, they can act in much more various situations.

In this study, therefore, we developed the hand which can be installed on life-size humanoid robot and which can support their whole-body. By using this hand, we achieved self-weight supporting motion. In this study, we used human-like musculoskeletal humanoid robot ``Kengoro''\cite{Kengoro_humanoids2016} as the robot on which the hand was installed.

In section I, we explained the background and problem of the motion for humanoid robots to support their body by hand comparing human. In section II, we will explain the specification of the hand which can support the weight of life-size humanoid robot. In section III, we will explain the design of the hand that we developed. In section IV, we will explain the experiments with humanoid robot ``Kengoro'' with the developed hand. In section V, we will state the conclusion of this study and future works.

\begin{figure}[t]
 \begin{center}
  \includegraphics[width=0.90\columnwidth]{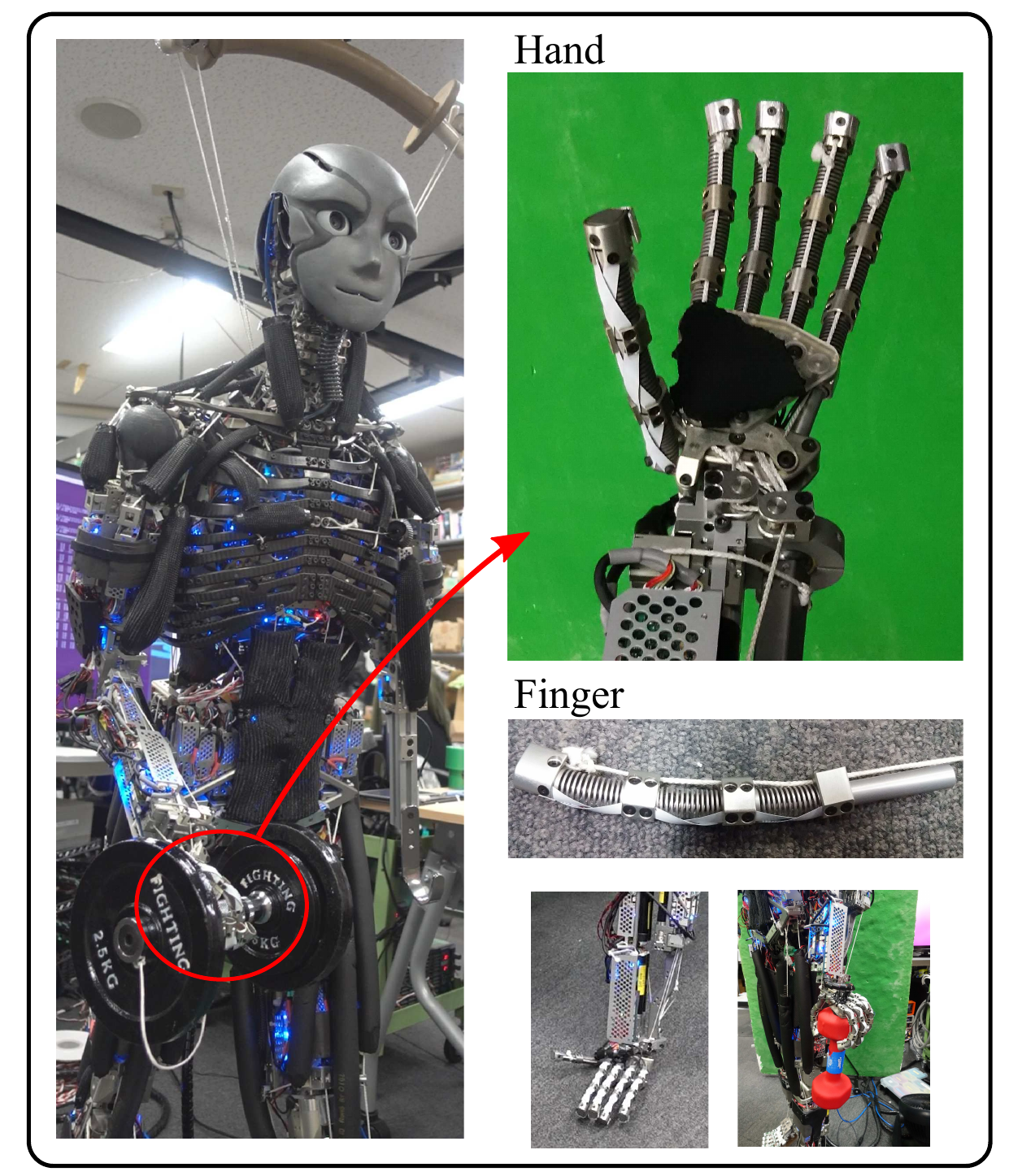}
  \caption{The developed hand with flexible fingers using machined springs and musculoskeletal humanoid Kengoro with the hand.}
  \label{figure:toppage}
  \vspace{-3.0ex}
 \end{center}
\end{figure}

\section{SPECIFICATION OF THE HAND}
We pointed out the following points for humanoid robots to support their support by hand.
\begin{enumerate}
 \item The hand has enough high power to support the weight of the humanoid robot.
 \item The hand has same proportion as human hand in the point of size and weight.
 \item The hand has enough robustness not to be broken by large force and impact.
 \item The hand has enough flexibility to change its shape along the object.
\end{enumerate}

Concerning 1), there are some motions that the body of humanoid robots must be supported only by hands without foots in the case or like that they dangle a bar. To do such a motion, the hand must have large gripping force. In this study, we aim for the hand to support the weight of ``Kengoro'', whose weight is 56.4[kg].

Concerning 2), some hands with large gripping force have been developed\cite{gifu_rsj2012}. But these hands were too large or too heavy to be installed on life-size humanoid robot. The hand also must be life-size for life-size humanoid to do the motion by using this.

Concerning 3), actuators need to exert the large force to get large gripping force. Moreover, the larger load the hand support, the larger impact it may take during motions. Therefore, the hand must have structural strength and impact resistance not to be broken.

Concerning 4), when humanoid robot do above-mentioned motions, they sometimes grasp the handle by hand and they at other times get their hands on the floor. The hand must manage both type of motions. In addition, it is desirable that many fingers bend along the objects to get large gripping force to various shapes of objects. In this way, the hand must change their shape to adapt to the scene.

To develop the hand which meets the above-mentioned specifications, we made wire-driven hand as human, which have palm and five fingers. To realize the enough strength to support whole-body, we used flexible joints in fingers using machined springs.

\begin{figure}[b]
 \begin{center}
  \includegraphics[width=0.99\columnwidth]{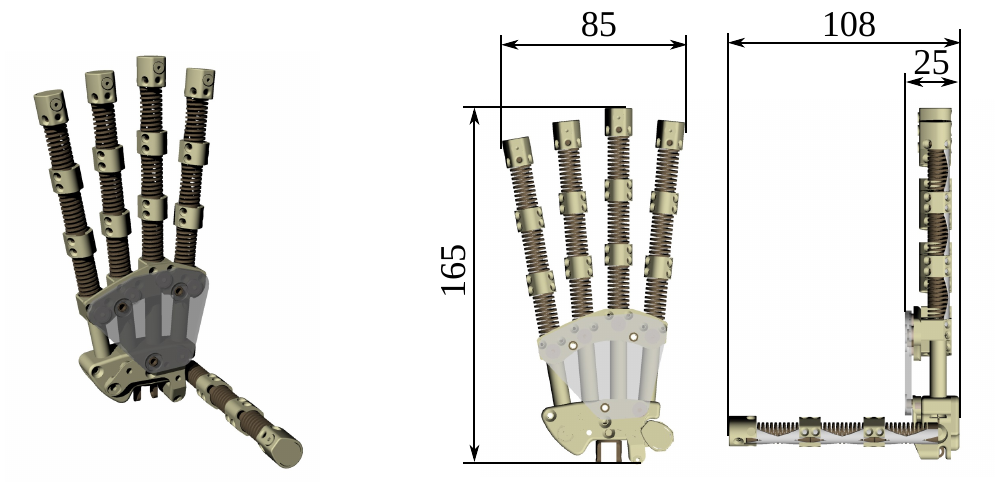}
  \caption{The size of the developed hand.}
  \label{figure:hand_all}
 \end{center}
\end{figure}

\begin{table}[b]
  \begin{center}
    \caption{Size and weight of human hand\cite{human_database_1997} and newly developed hand.}
    \label{table:hand_size}
    \begin{tabular}{|c|c|cc|}
      \hline
      \multicolumn{2}{|c|}{}& Human & This hand\\
      \hline
      Weight & Hand  & 0.6 & 0.3\\
      {\textrm [}kg{\textrm ]} & Forearm \cite{Kawaharazuka_iros2017} & 1.2 & 1.5\\
      \hline
      Size & Length & 180.6 & 165\\
      {\textrm [}mm{\textrm ]} & Breadth & 81.8 & 85\\
      & Thickness & 28.9 & 25\\
      \hline
    \end{tabular}
  \end{center}
\end{table}

\section{DESIGN OF THE HAND}
\subsection{Overview}

The five-finger hand we designed is shown in \figref{hand_all}. This is wire-driven and underactuated hand. As with extrinsic muscle, one of the structure of human, the actuators to tighten wires are arranged in forearm\cite{Kawaharazuka_iros2017}, where there is more space than in hand and wires extended from actuators are connected fingertip by way of wrist. This enables us to use relatively high-power 60W actuators for hand which can exert large wire tension of 50[kgf].

The size and weight of this hand are shown in \figref{hand_all} and \tabref{hand_size} respectively. These can be considered human-like. Although the forearm is a little heavier than that of human because of the weight of actuators and boards for actuators and sensors, we made the hand a little lighter, so that we kept human-like weight as a whole.

The skeletal structure also imitates human. Five fingers extend from the parts of carpal near the wrist. Thumb is placed as it oppose four fingers so as to make it easier to grasp objects. But the palm, nonetheless, can contact to the surface because the thumb can be flexed by reaction force from the ground, wall, and others. Palm consists of the polycarbonate plate which is placed on the metacarpal bones, which are the most proximal bones of fingers. Wrist joint consists of universal joint so that it has two DOFs: abduction and adduction, flexion and extension. The wrist joint can be small because wires as well as the joint support the load. Thus, the size as with human can be achieved.

\subsection{Fingers Using Machined Springs}

\begin{figure}[t]
 \begin{center}
  \includegraphics[width=0.40\columnwidth]{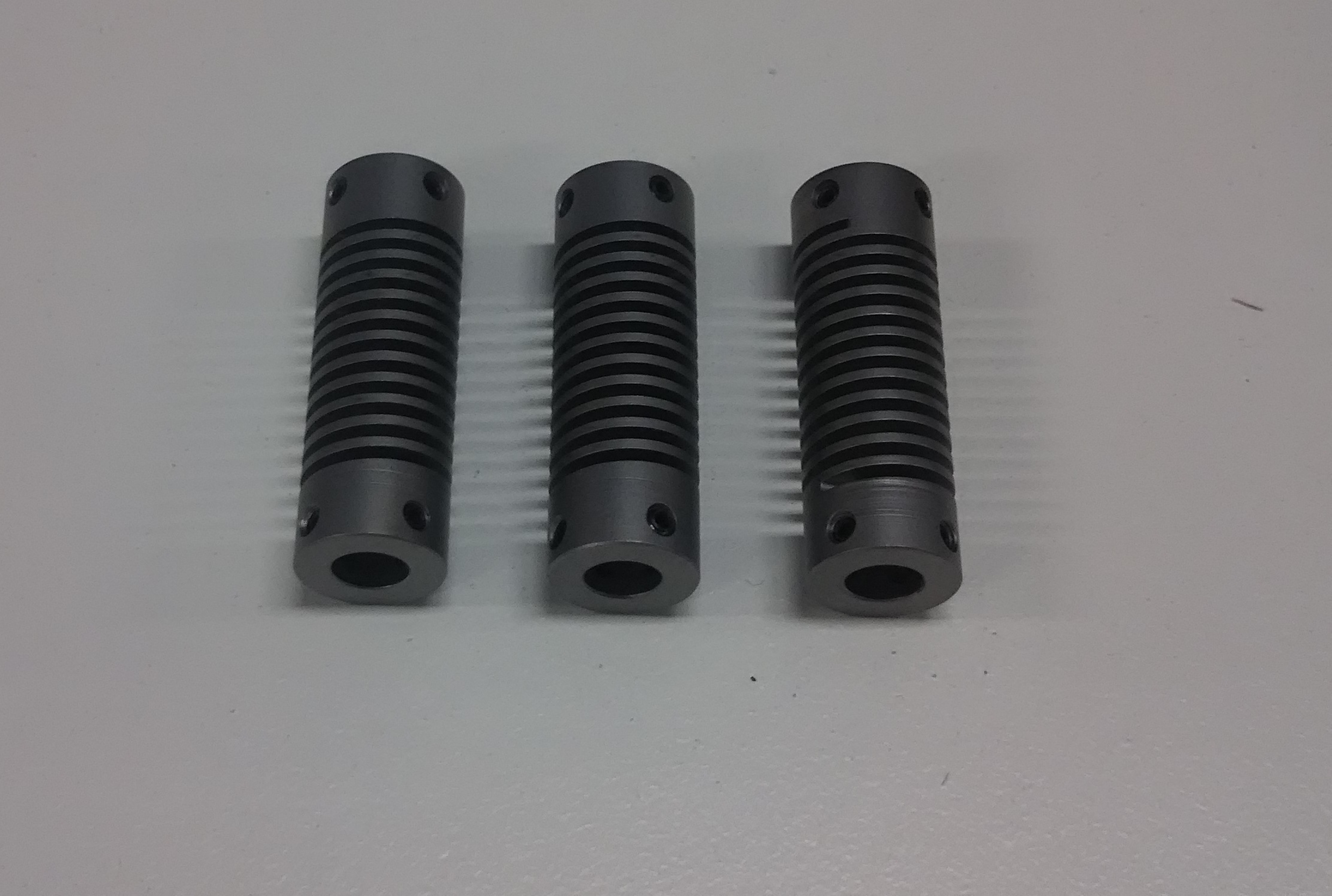}
  \includegraphics[width=0.50\columnwidth]{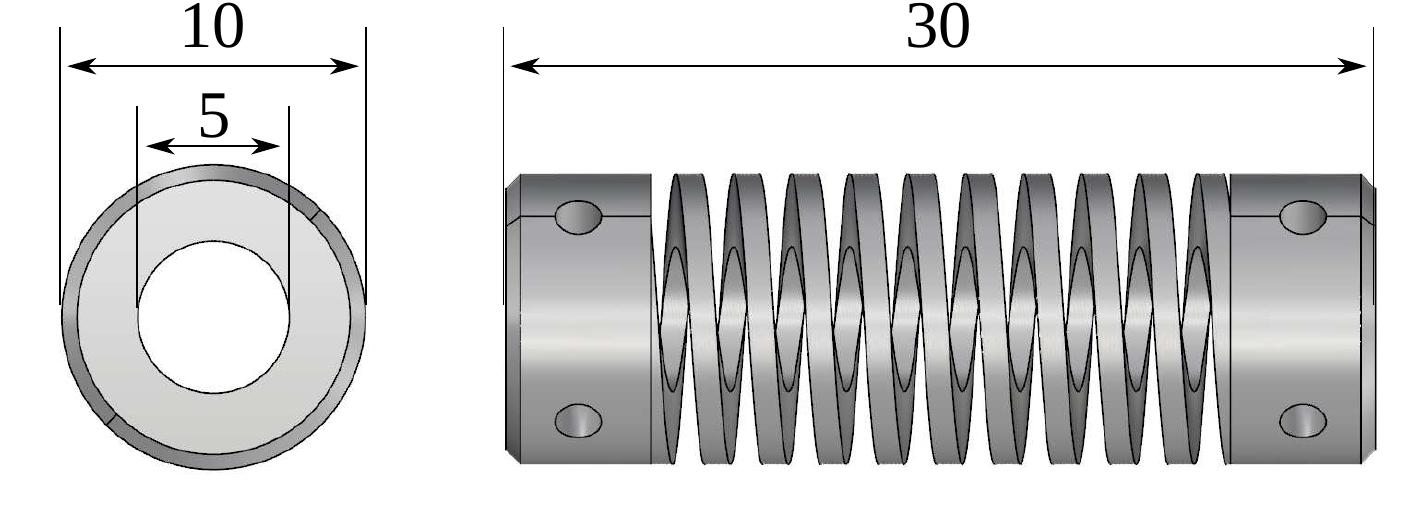}
  \caption{Machined Spring.}
  \label{figure:spring}
 \end{center}
\end{figure}

\begin{table}[t]
  \begin{center}
    \caption{Spring constants of machined springs in each joint[deg/Nm].}
    \label{table:spring_stiffness}
    \begin{tabular}{|c|c|c|c|}
      \hline
      & CM Joint & MP Joint & IP Joint\\
      \hline
      Thumb & 664 & 443 & 443\\
      \hline
      \hline
      & MP Joint & PIP Joint & DIP Joint\\
      \hline
      Index & 863 & 664 & 443\\
      Middle & 903 & 707 & 443\\
      Ring & 863 & 664 & 443\\
      Little & 707 & 664 & 443\\
      \hline
    \end{tabular}
  \end{center}
\end{table}

As the structures of joints which are often used in hands, axis joints\cite{shadow_industrialrobot}, rubber joints\cite{Carrozza_icra2005} and coil spring joints\cite{iowa_industrialrobot} can be considered. However, these structures have weakness in the points of little and thin finger for self-weight supporting motion of humanoid robots. Axis joints are weak in shock. Rubber joints are broken off by large force though they can turn shock aside. General coil spring joints are so difficult to connect other parts that the hand can be structurally weak.

Therefore, we use machined springs as joints. Machined spring literally means the spring made by machining. Machined spring has some merits. For example, it is used to simplify small manipulator for medical use\cite{Haraguchi_iros2011}. In this paper, we use it to make strong finger.

In the first place, the following points are considered as the merit of using spring joint.
\begin{itemize}
\item It is strong because of metal and can turn shock aside because of elasticity.
\item Restoring force can be used in order for fingers to extend without additional spring or wire.
\end{itemize}
Next, the following points are considered as the merit of using machined spring joints compared to general coil spring joint made by bending.
\begin{itemize}
\item It can be connected firmly to other parts because the spring and the attachments can be integrally made as one piece.
\item The cross-sectional area can be larger for the size of spring because it can be quadrangle. Thus, it is advantageous to space-saving in the same strength.
\end{itemize}
By using machined springs which have above-mentioned merits, the finger joints can be made which is stronger in shock than axis joints, which can withstand larger force than rubber joints, and in which parts to each other are firmly connected.

\begin{figure}[t]
 \begin{center}
  \includegraphics[width=0.95\columnwidth]{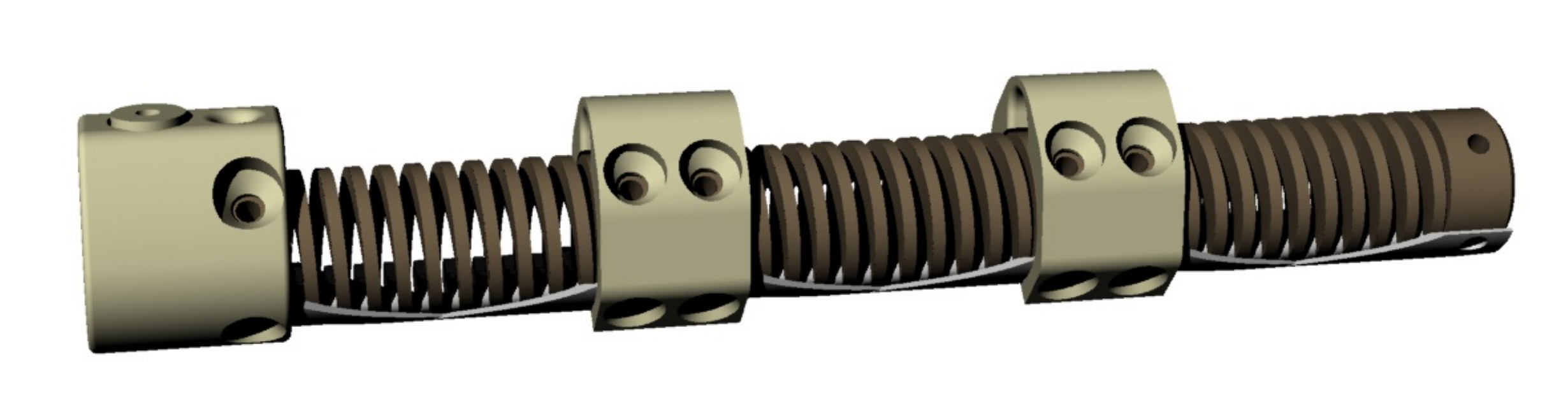}
  \includegraphics[width=0.99\columnwidth]{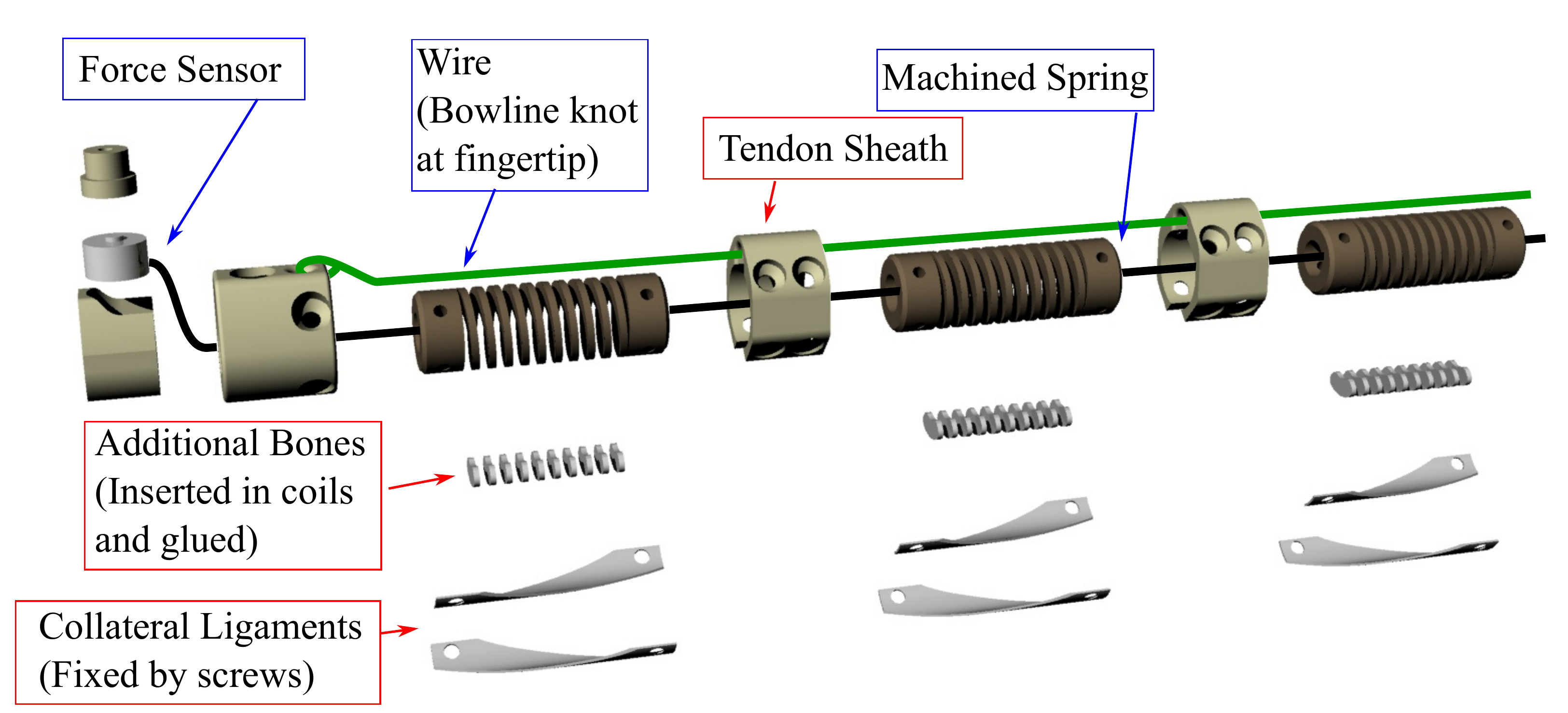}
  \caption{The finger using machined springs.}
  \label{figure:finger_all}
 \end{center}
\end{figure}

In this hand, we used machined springs made of stainless steel (SUS 630). The finger is shown in \figref{finger_all}. One finger with three joints consists of three machined springs and connection parts which imitate tendon sheath. This three joints correspond to CM, MP and IP joints in thumb or MP, PIP and DIP joints in other fingers. For fingers to flex along objects in grasping, the most proximal spring is the softest in each finger and the most distal is the stiffest. The spring constants of the springs in this hand are shown in \tabref{spring_stiffness}. One wire is used to flex a finger. The finger is flexed when the wire is tightened and extended by restoring force of springs when the wire is loosened.

However, in addition to flexion and extension, one more rotational DOF occurs only by the connection of springs: abduction and adduction. And what is worse, the length of springs can be shrunk and elongated. To inhibit these extra DOFs, steel thin plates and small plastic parts are placed. Steel plates imitate ligaments. They inhibit extra elongation of length and rotation of abduction and adduction. Plastic plates are inserted between coil of springs and glued. This plates reproduce geometric constrains of contact between the bones. These plates support the certain motion of fingers. However, the DOF of abduction and adduction is left a little rather than completely inhibited in the point of shock-absorbing.

\begin{figure}[tbh]
 \begin{center}
  \includegraphics[width=0.98\columnwidth]{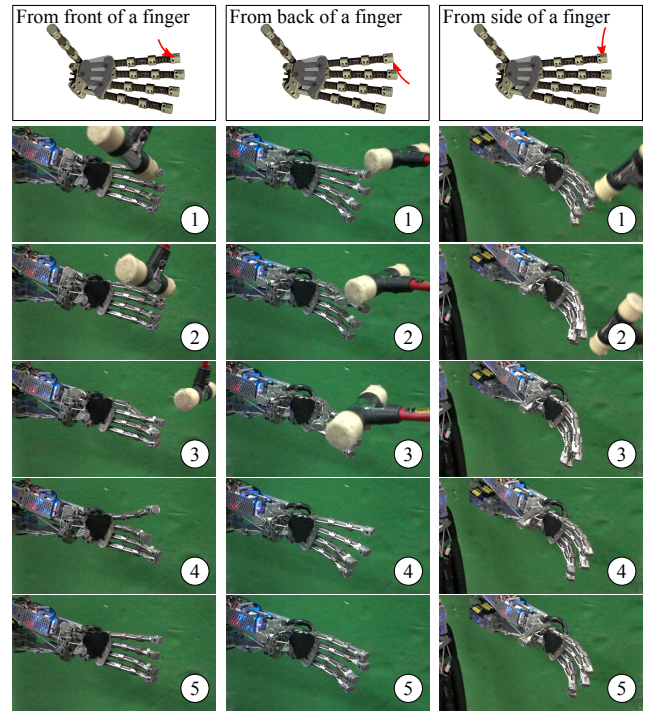}
  \caption{Experiment of adding shock to fingers with a hammer.}
  \label{figure:hammer}
 \end{center}
\end{figure}

As the experiment to confirm the strength in shock, we added shock to fingers with a hammer. We hit the fingertip of index finger from the front and the back of the finger  when fingers were extended. In addition, we hit the thumb from the side in order to confirm the strength in shock of abductional and adductional direction. A sequence of pictures in this experiment is shown in \figref{hammer}. In the same way, we also experimented when the joints were flexed to 30[deg] in each joint. In all state, fingers could turn shock aside without broken and were returned to the original position because they have elasticity.

\subsection{Other Characteristics}
\subsubsection{Arrangement of Wires}
\begin{figure}[t]
  \begin{minipage}{0.46\linewidth}
    \begin{center}
      \includegraphics[width=0.98\columnwidth]{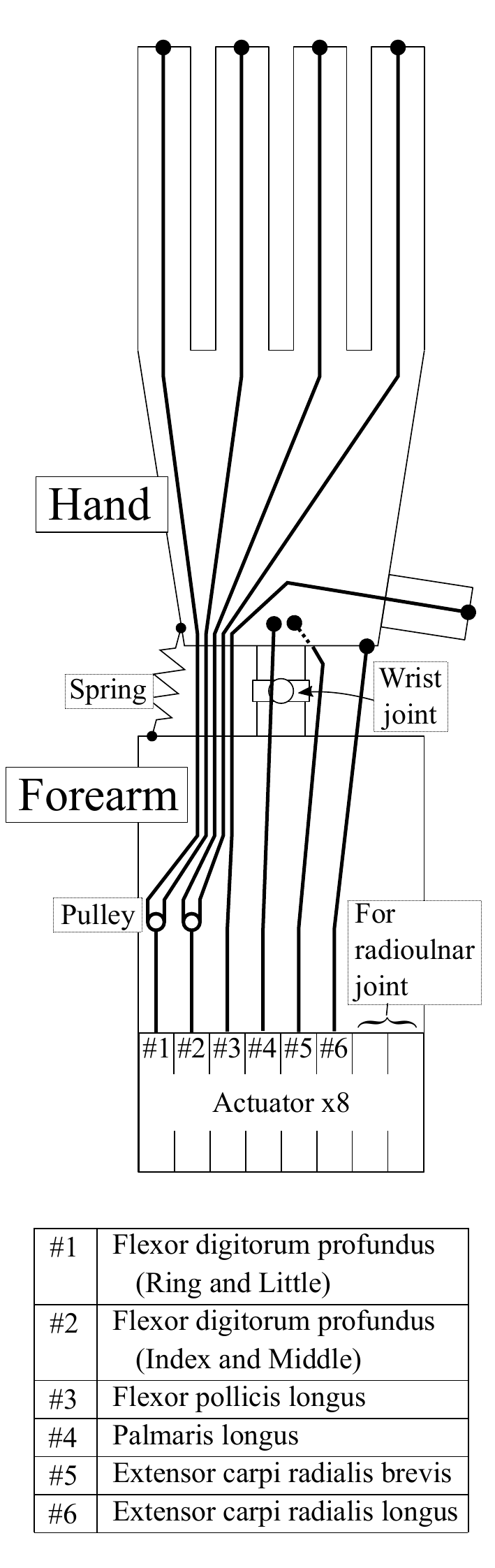}
      \caption{Schematic view of wire path.}
      \label{figure:wire_arrangement}
    \end{center}
    \begin{center}
      \includegraphics[width=0.98\columnwidth]{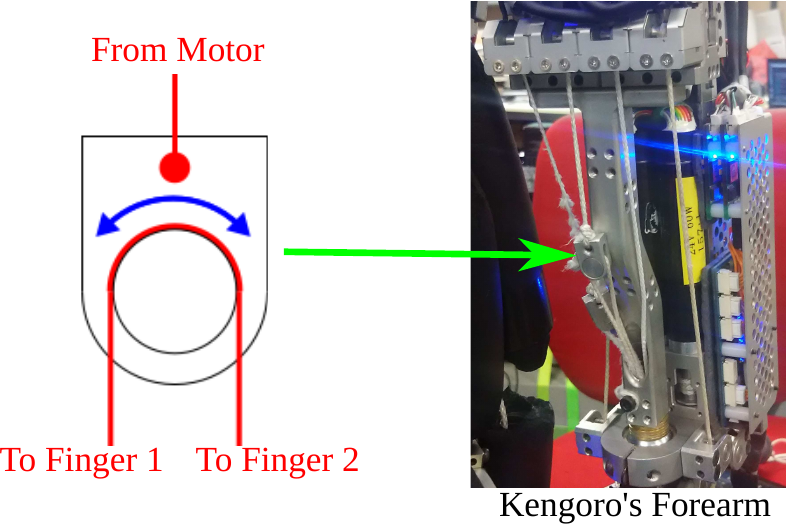}
      \caption{Movable pulley.}
      \label{figure:movable_pulley}
    \end{center}
  \end{minipage}
  \begin{minipage}{0.50\linewidth}
    \begin{center}
      \includegraphics[width=0.98\columnwidth]{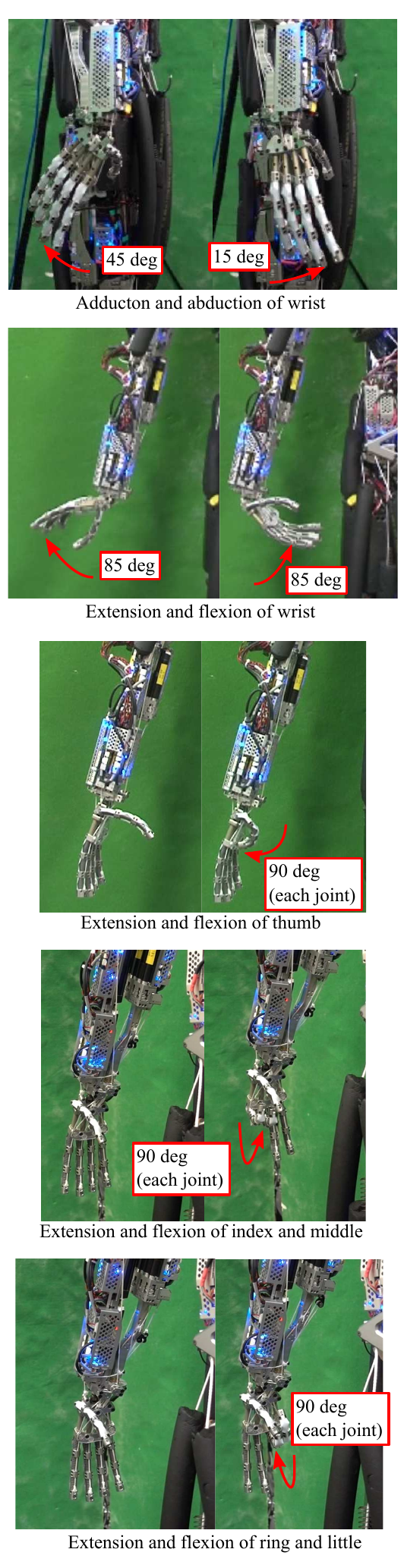}
      \caption{Movement of each joint.}
      \label{figure:hand_basic_move}
    \end{center}
  \end{minipage}
\end{figure}

\begin{figure}[t]
 \begin{center}
  \includegraphics[width=0.7\columnwidth]{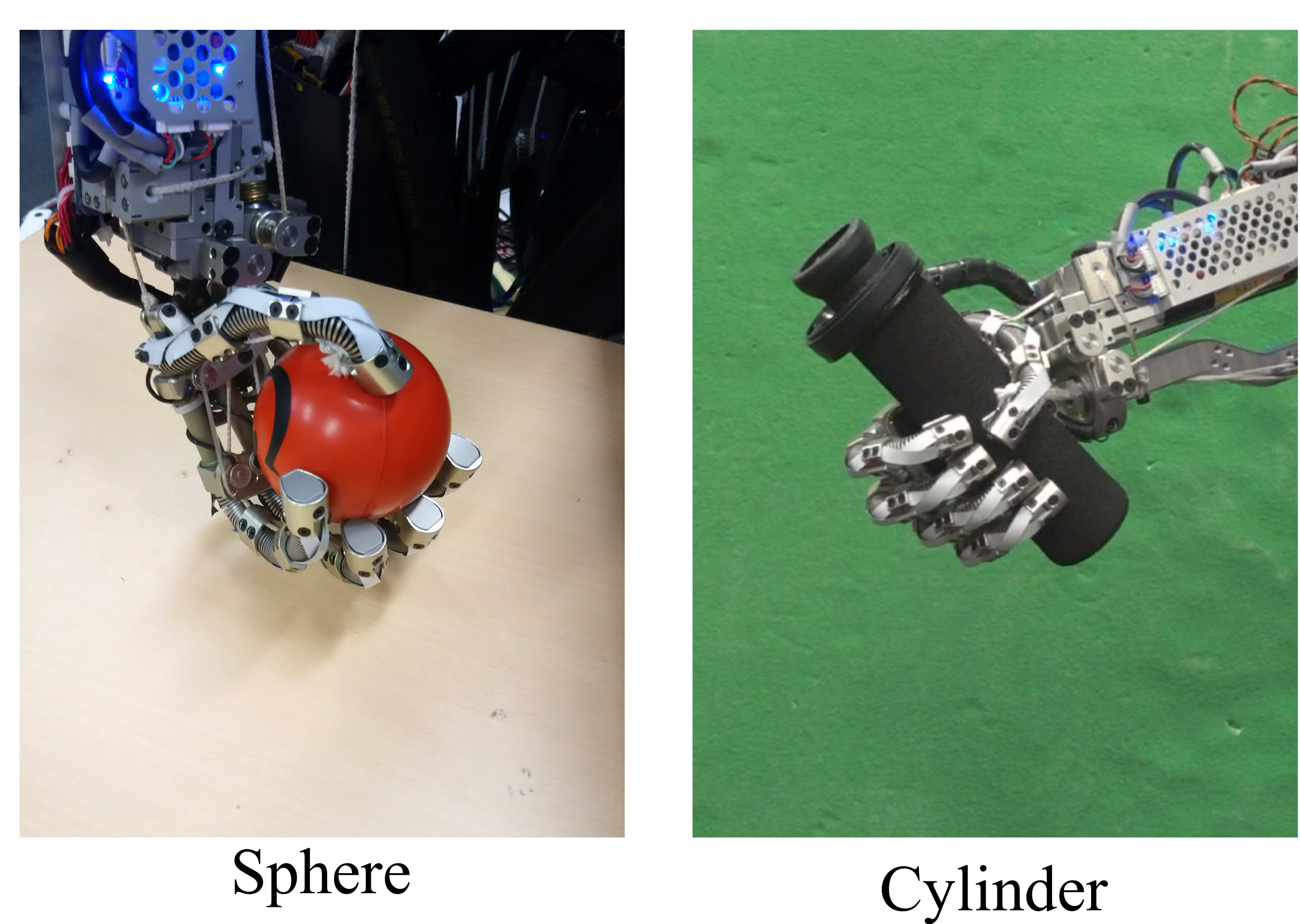}
  \caption{Grasp of sphere and cylinder.}
  \label{figure:finger_bend}
 \end{center}
\end{figure}

8 actuators are placed in forearm, all of which can control the length and tension of wires\cite{Kawaharazuka_iros2017}. Because two of them are used for pronation and supination of radioulnar joint, the others are used for wrist and fingers. Wrist joints are moved by three wires and one coil spring. Five fingers are moved by the other three wires. Schematic view of wire path and the list of muscles corresponding to wires are shown in \figref{wire_arrangement}. The pair of index and middle finger, and that of ring and little finger is moved by same actuator because the number of wire is less than that of finger. Thus, we inserted the movable pulley shown in \figref{movable_pulley} between actuator and fingertips. It is used in some studies\cite{movable_pulley_icra2008}. Therefore, each finger can be flexed along objects even if one finger touch object earlier than other fingers. The movement of each joint is shown in \figref{hand_basic_move}. The grasp of various shape of object using the characteristics of discrete flexion of each finger is shown in \figref{finger_bend}.

\subsubsection{Arrangement of Force Sensors}
\begin{figure}[t]
 \begin{center}
   \includegraphics[width=0.40\columnwidth]{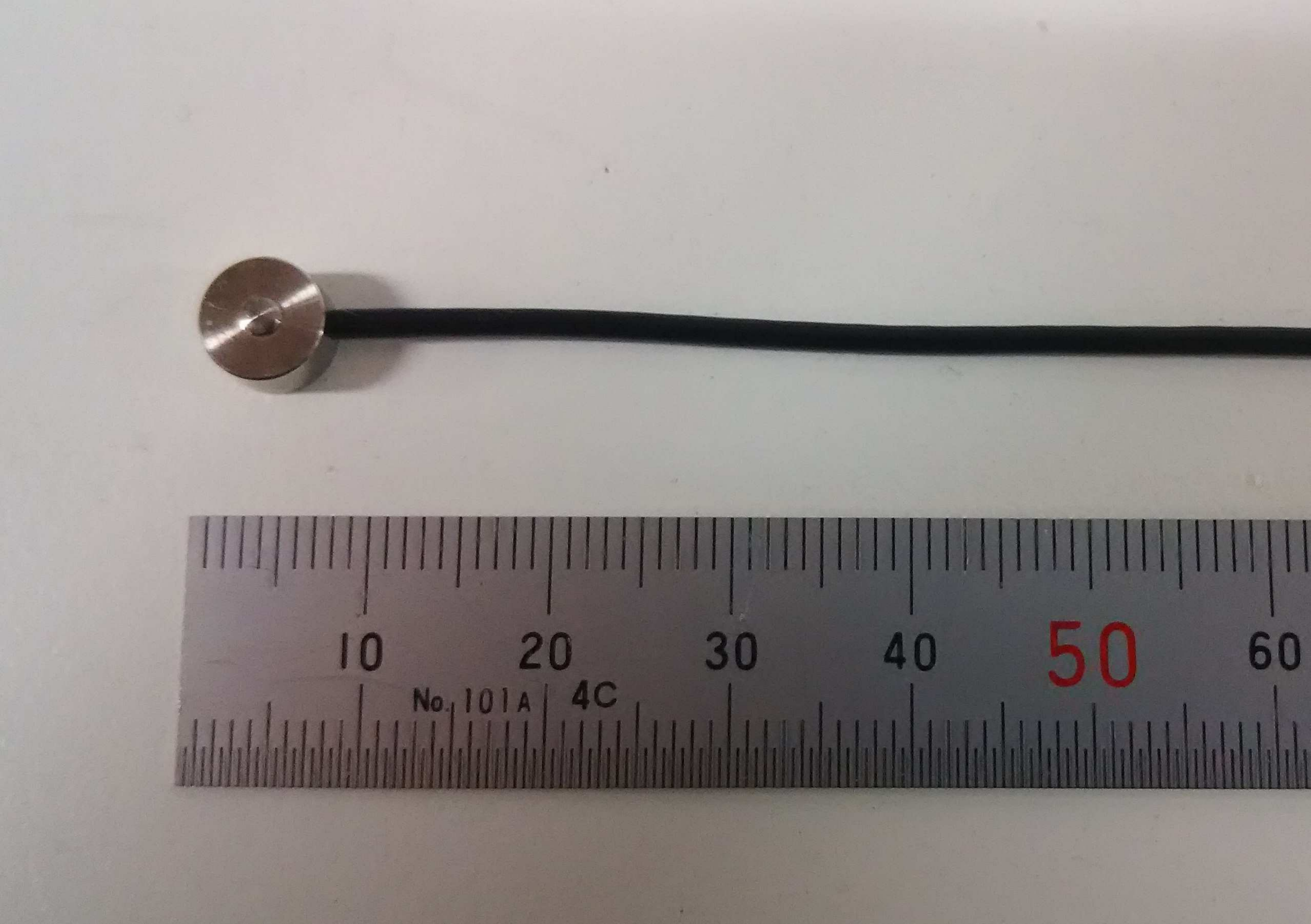}
   \includegraphics[width=0.35\columnwidth]{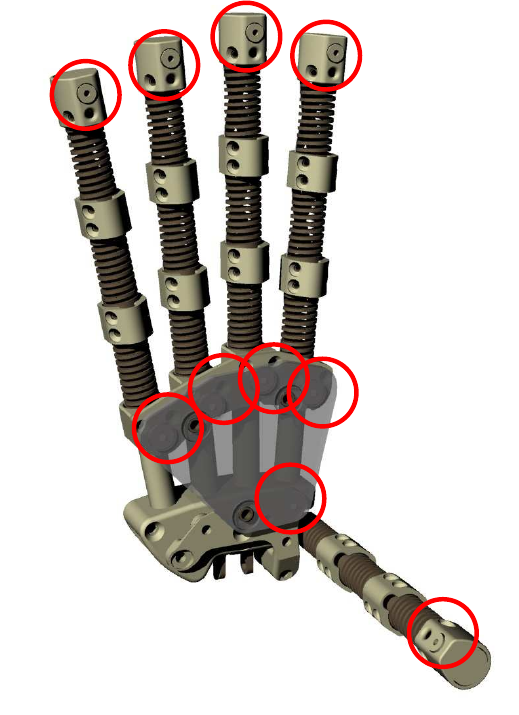}
  \caption{Arrangement of force sensors.}
  \label{figure:loadcell_arrangement}
 \end{center}
\end{figure}

\begin{figure}[tbh]
 \begin{center}
  \includegraphics[width=0.99\columnwidth]{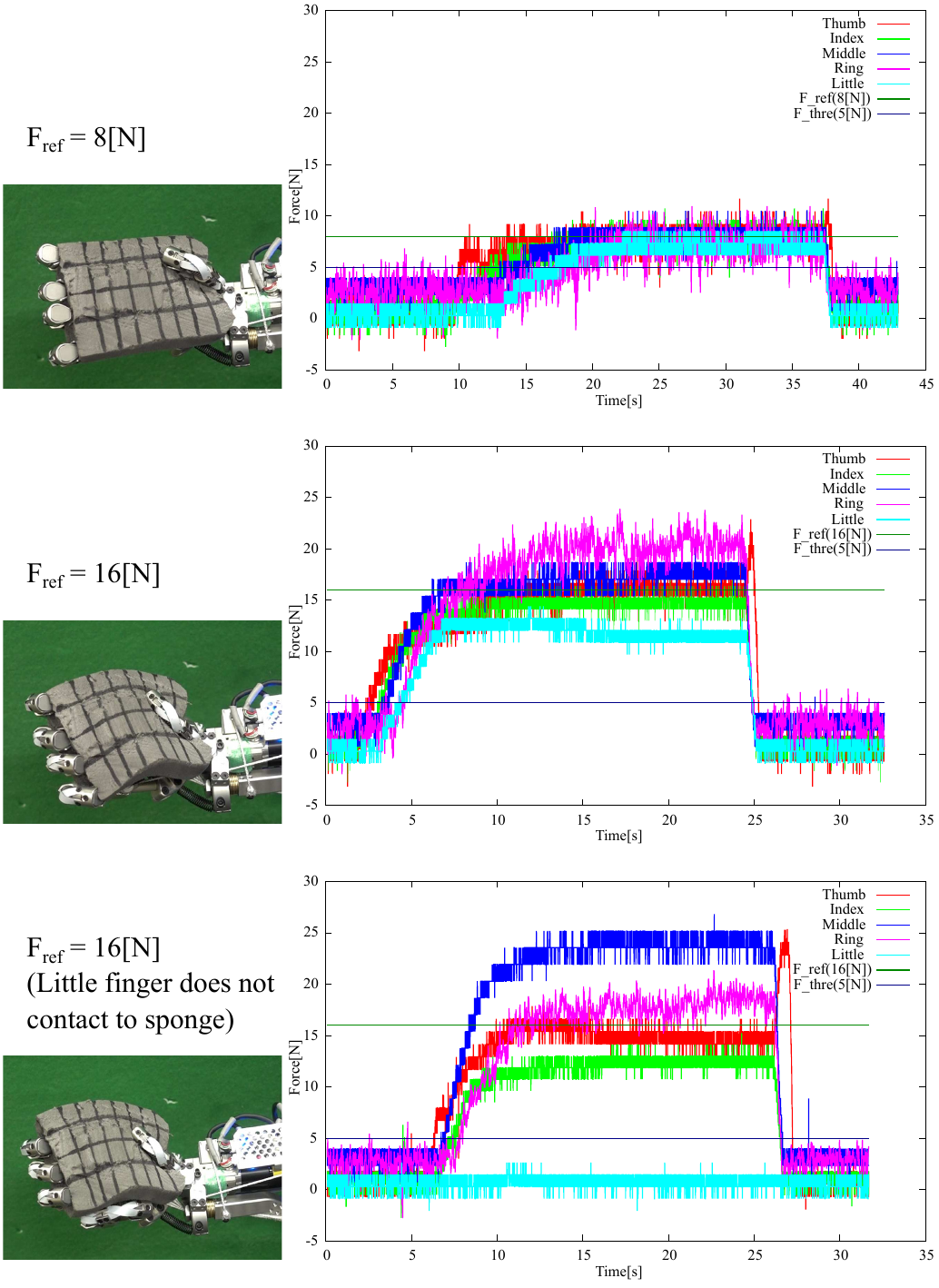}
  \vspace{-3.0ex}
  \caption{Experiment of grasping sponge.}
  \label{figure:sponge_grasp}
 \end{center}
\end{figure}

In this hand, 10 force sensors (MCDW-50L: 500[N] max.) are placed in fingertips of each finger and palm to measure external force (\figref{loadcell_arrangement}). These sensors can be available to detect grasping object, keep gripping force constant and others. As an example of grasp motion with force sensors, we had this hand grasp the soft sponge with small gripping force and large gripping force (\figref{sponge_grasp}).

Basically, the length of each wire is controlled depending on the value of force sensor in fingertip which the wire bends ($F_{finger}$ in Eq. 1). In thumb, $F_{finger}$ is the sensor value in thumb fingertip. In four fingers of this hand, the average of the values of two sensors ($F_1$, $F_2$) is used as $F_{finger}$ because the wire from one actuator branches at the pulley and bends two fingers. Thus, one finger applies two times as large force as $F_{ref}$ when the other finger does not contact to the object. To prevent it, the threshold value is used to judge whether each fingertip contact to object. Only one value of the two sensor value is referred rather than average value if only one of two fingertips contacts (ring and little, in the bottom of \figref{sponge_grasp}). To summarize the above, the equation to adjust wire length are shown in Eq. 1 and Eq. 2.

But there is still a problem. One finger applies larger force than $F_{ref}$ when the other cannot apply as large force as $F_{ref}$. It occurs in the case or like that the force is less but not zero when it have been flexed to the edge of the motion range of finger joints. In the bottom of \figref{sponge_grasp} the sensor value in middle is larger than $F_{ref}$ because that in index is smaller. Therefore, the improvement of the control is required. In addition, the control using not only the sensors in fingertips but also those in palm, which are not used in this experiment, will be implemented.

\begin{equation}
\Delta l = k \left(F_{finger} - F_{ref}\right)
\end{equation}
\begin{equation}
F_{finger} = \left \{ \begin{array}{cl}
F_{thumb} & \left(Thumb\right)\\
\displaystyle\frac{F_1 + F_2}{2} &
\left(\left(F_1 \geq F_{thre}, F_2 \geq F_{thre}\right)\right.\\
&or\,\left.\left(F_1 < F_{thre}, F_2 < F_{thre}\right)\right)\\
F_1 & \left(F_1 \geq F_{thre}, F_2 < F_{thre}\right) \\
F_2 & \left(F_1 < F_{thre}, F_2 \geq F_{thre}\right) \\
\end{array} \right.
\end{equation}
$where$\\
$l$ : Wire length.\\
$F_{ref}$ : Reference of force sensor value.\\
$F_{thumb}$ : The value of the force sensor in the fingertip of the thumb.\\
$F_1, F_2$ : Values of force sensors in the fingertip to which the wire is connected.\\
$F_{thre}$ : Threshold value of judging contact (In this experiment, $F_{thre} = 5[{\mathrm N}]$).\\


\section{SELF-WEIGHT SUPPORTING MOTION WITH HAND}
\subsection{Grasp of Weights in Basket}
To confirm if the developed hand has enough gripping force, we made an experiment that we added weight to basket grasped by the hand.

\begin{figure}[tbh]
 \begin{center}
  \includegraphics[width=0.60\columnwidth]{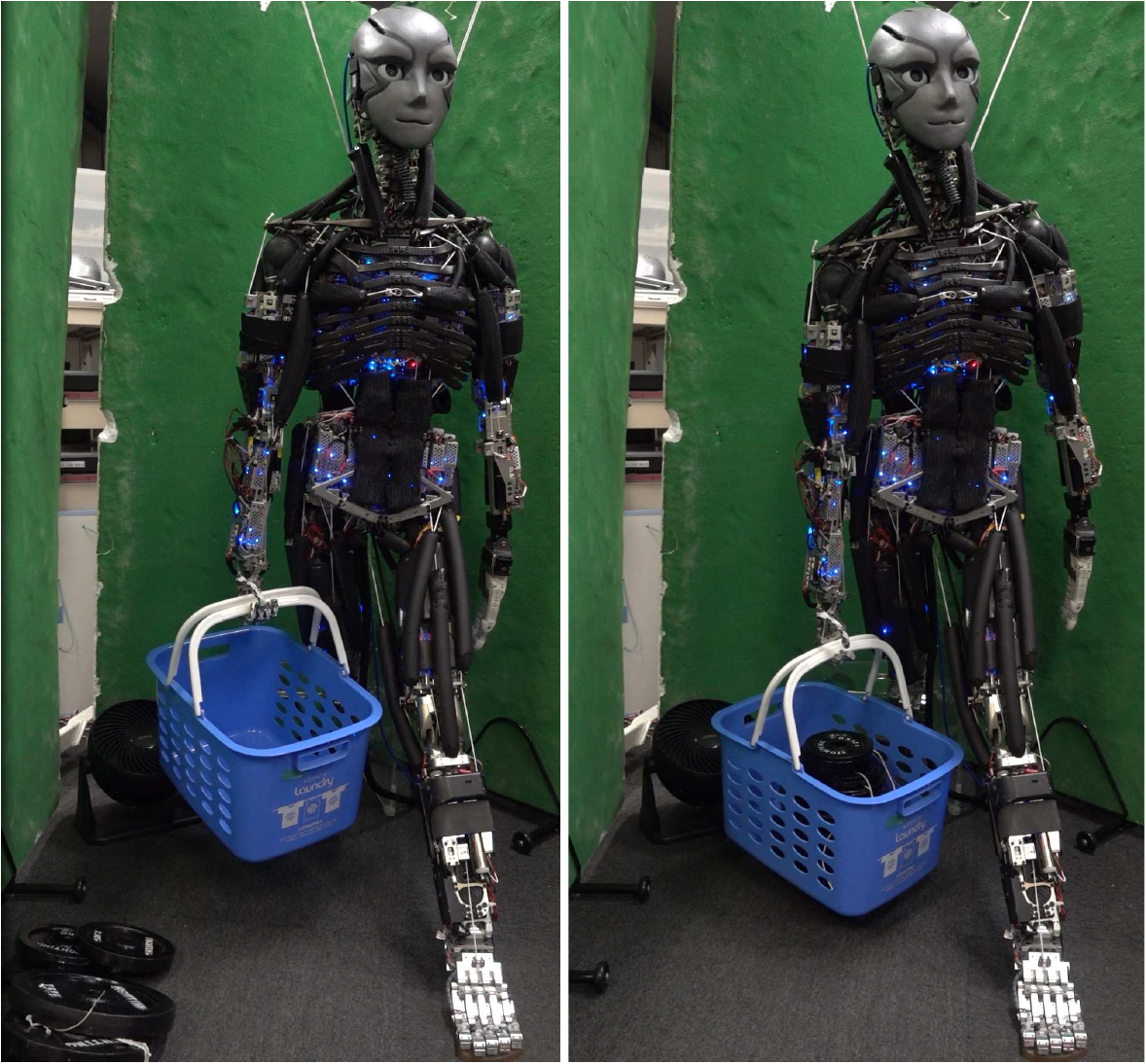}
  \vspace{-1.0ex}
  \caption{Experiment of grasping weight in basket.}
  \label{figure:omori}
 \end{center}
\end{figure}

\begin{figure}[tbh]
 \begin{center}
  \includegraphics[width=0.95\columnwidth]{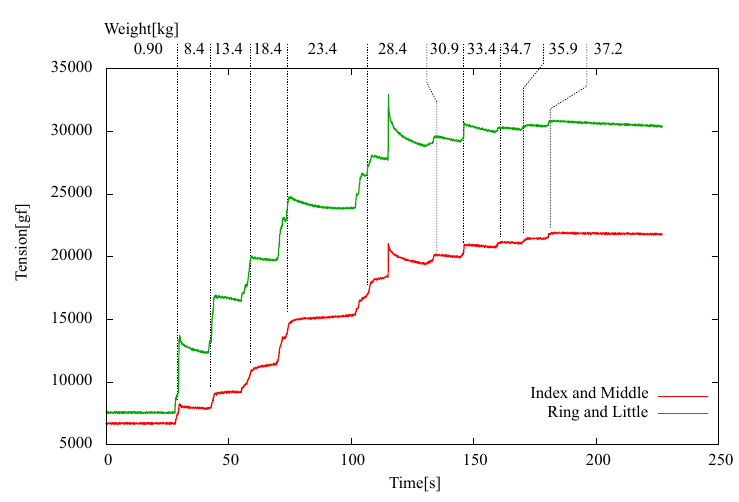}
  \vspace{-1.0ex}
  \caption{Change of wire tension.}
  \label{figure:omori_graph}
  \vspace{-3.0ex}
 \end{center}
\end{figure}

First, we had the hand grasp the empty basket (0.9[kg]). After this, grasping state is kept by the control keeping the length of wires. Next, we gradually added weight to the basket. As the result, the hand kept grasping the basket without dropping even when the total weight of basket and weight became 37.2[kg] (\figref{omori}). The change of wire tension to flex fingers is shown in \figref{omori_graph}. Maximum wire tension, when the weight is the heaviest, is about 30.0[kgf]. This hand can support with an margin about half weight of ``Kengoro'' (56.2[kg]), considering the maximum wire tension which the actuator can exert is about 50.0[kgf].

\subsection{Push-up Motion}
\begin{figure}[t]
 \begin{center}
  \includegraphics[width=0.95\columnwidth]{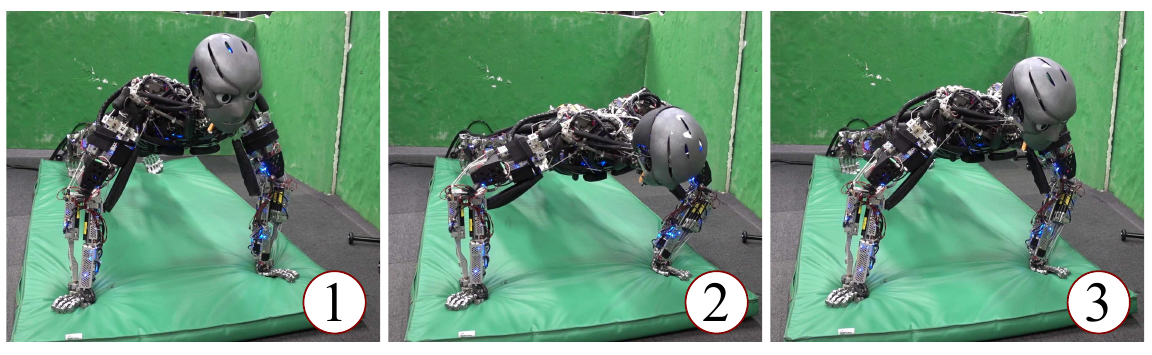}
  \caption{Push-up motion.}
  \label{figure:udetate}
 \end{center}
\end{figure}

We had Kengoro do push-ups as the motion to support its body by palm. In this experiment, four limbs supported the whole body. The thumb are flexed and placed next to the palm when the palms contact to the floor. This and the flexion of wrist joints along the floor enabled the hand to support the body stably. A sequence of pictures in this experiment is shown in \figref{udetate}.

\subsection{Dangling Motion}
\begin{figure}[tbh]
 \begin{center}
  \includegraphics[width=0.95\columnwidth]{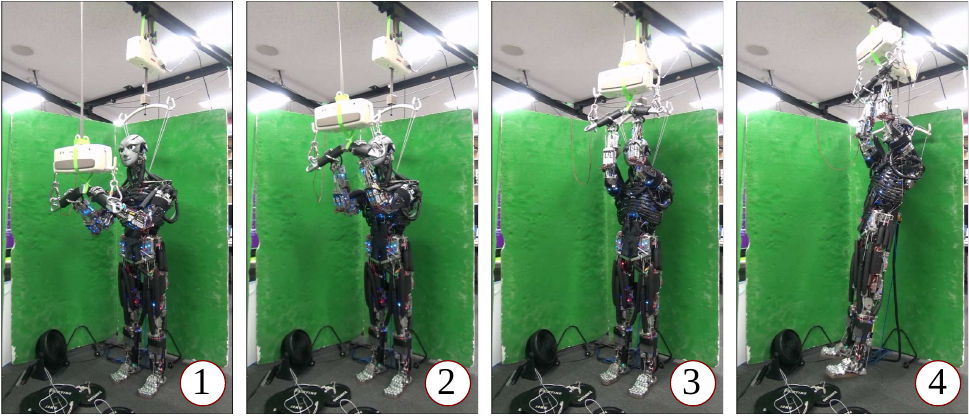}
  \vspace{-1.0ex}
  \caption{Dangling motion.}
  \label{figure:burasagari}
  \vspace{-3.0ex}
 \end{center}
\end{figure}

\begin{figure}[tbh]
 \begin{center}
  \includegraphics[width=0.95\columnwidth]{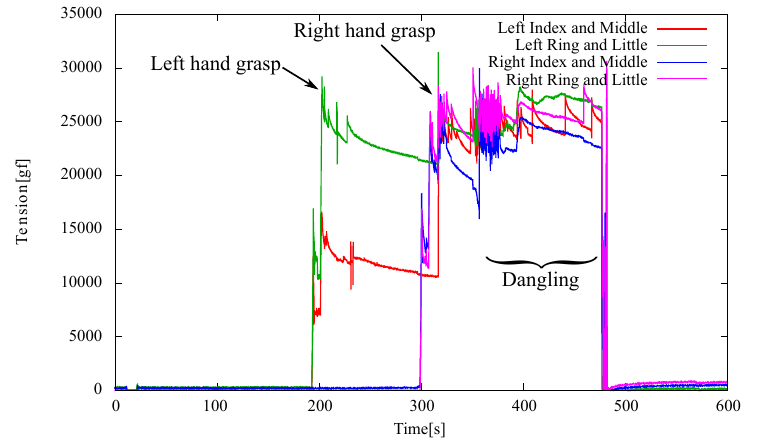}
  \vspace{-1.0ex}
  \caption{Wire tension during dangling motion.}
  \label{figure:grasp_with_ar_tension}
  \vspace{-3.0ex}
 \end{center}
\end{figure}

We had Kengoro dangle a bar as the motion to support its body by grasping. First, Kengoro grasp a bar in front of its shoulder. Second, the state that finger was bend was kept by the control of which keeps wire tension constant. Third, the bar was lifted by a crane. As the result, its body could be supported only by hands. A sequence of pictures in this experiment is shown in \figref{burasagari}. The maximum wire tensions are also about 30.0[kg] in this experiment  as shown in \figref{grasp_with_ar_tension}. Dangling motion could be achieved within the range of wire tension that actuators can exert.

\section{CONCLUSION}
In this study, we developed the hand which can be installed on life-sized humanoid and which has enough gripping force to support whole body. We achieved the characteristic of three points by using machined springs and some additional plates.
\begin{itemize}
  \item the size as with human hand
  \item the structural strength not to be broken by large load and high wire tension
  \item the robustness with which impact it may takes during motions can be turned aside
\end{itemize}

In addition, fingers can be flexed along objects by movable pulley. There are planes to support the robot. Large gripping force is produced by relatively more high-power actuators than ordinary placed in forearms. We had musculoskeletal humanoid ``Kengoro'' do the motions in which it supports its body by hand: push-up motion and dangling motion and showed the usefulness the developed hand.

As the future work, we would like to achieve more various motions with this hand. In order to do that, it is necessary for the control of limbs as well as hands of tendon-driven humanoid to be improved. And we also would like to develop more high-performance hand with dexterity to be able to apply the large gripping force to more various shapes of objects by improving bone structure, joint DOFs, wire paths and others.
\addtolength{\textheight}{-20cm}   









\bibliographystyle{IEEEtran}
\bibliography{string, iros2017-makino, common}

\end{document}